# Fast Skill Learning for Variable Compliance Robotic Assembly

Tianyu Ren
ultrain@126.com
State Key Laboratory of Tribology, Department of Mechanical Engineering, Tsinghua University, Beijing 100084, China

Yunfei Dong
d_yunfei@163.com
State Key Laboratory of Tribology, Department of Mechanical Engineering, Tsinghua University, Beijing 100084, China

Dan Wu*
wud@tsinghua.edu.cn
State Key Laboratory of Tribology, Department of Mechanical Engineering, Tsinghua University, Beijing 100084, China

Ken Chen
kenchen@tsinghua.edu.cn
State Key Laboratory of Tribology, Department of Mechanical Engineering, Tsinghua University, Beijing 100084, China

*Abstract*
The robotic assembly represents a group of benchmark problems for reinforcement learning and variable compliance control that features sophisticated contact manipulation. One of the key challenges in applying reinforcement learning to physical robot is the sample complexity, the requirement of large amounts of experience for learning. We mitigate this sample complexity problem by incorporating an iteratively refitted model into the learning process through model-guided exploration. Yet, fitting a local model of the physical environment is of major difficulties. In this work, a Kalman filter is used to combine the adaptive linear dynamics with a coarse prior model from analytical description, and proves to give more accurate predictions than the existing method. Experimental results show that the proposed model fitting strategy can be incorporated into a model predictive controller to generate good exploration behaviors for learning acceleration, while preserving the benefits of model-free reinforcement learning for uncertain environments. In addition to the sample complexity, the inevitable robot overloaded during operation also tends to limit the learning efficiency. To address this problem, we present a method to restrict the largest possible potential energy in the compliance control system and therefore keep the contact force within the legitimate range.

*Note to Practitioners*
Assembly is a labor-intensive work in manufacturing industries where automation is highly needed. Though the combination of deep reinforcement learning and variable-compliance action of robot has been shown significant robustness and adaptability to environmental change and disturbance compared with constant stiffness strategies (analog to RCC device in robotic assembly), the acquirement of such policy remains difficult even with the remarkable progress taking place in machine learning society. For skill learning of physical robotic system, the learning speed affects directly the production efficiency so that it should be carefully addressed to ensure the applicability of the learning-based method for industrial practitioners. However, the reinforcement learning in physical world reveals different difficulties, mainly including the sampling efficiency and the exploration safety. In this work, we propose a new data fusion algorithm to fit local model for efficient model-guided exploration of the variable compliance policies. In addition, to ensure a safe exploration and reduce the training assistance time, a contact force restriction controller is designed and activated when the robot exceeds a pre-defined safe region. Experimental results demonstrate an improvement of policy learning speed in robotic assembly with the proposed method. Currently the prior model in local model fitting is given by human in analytic form. Although this expression is quite simple and intuitive, we hope to replace it with human demonstration to further improve the accessibility.

*Paper type* Research paper

## I. INTRODUCTION

In robot manipulation, the peg-in-hole problem represents a group of contact-rich tasks with minor model information. Though the success of deep reinforce learning has enable task execution without analyzing the detail physical model, the robot operation has to be restricted carefully to low speed to avoid potential overload or impact[1]. On the other hand, it is helpful to adapt the ideas underlying the success of human work: variable compliance control and learning. The success of remote center of compliance (RCC) device has shown the importance of compliance regulation in insertion tasks [2]. Yun [3] further concluded that the passive compliance can greatly help with the peg-in-hole task and it yields more stability in contact motion. On the other hand, the emerging collaborative robots with torque-controlled joints show their ability of programmable compliance without auxiliary device and therefore give a much more flexibility for various tasks [4, 5]. However, finding the appropriate compliance policy in a short time is usually a hard problem [6].

Reinforcement learning (RL) [7] is recognized as an approach used by human to learn responses to situations through associating corresponding rewards [8]. It is a possible solution to problem of optimal control without knowledge of the models of the robot &environment system [9]. Combining RL with path integrals (PI2), variable stiffness control of robot has been accomplished [10, 11]. The strategies of both reference trajectory and the stiffness of the end-effector are generated by the planner from a stochastic optimal process with path integrals, and they are successfully implemented in via-point tracking, pool stroke [12] and pancake-flipping [13]. However, for contact-rich tasks such as robotic assembly, the robot have to gather large amounts of experience before finding an effective policy for the complex control tasks. The implementation of the variable compliance controller inevitably introduces more correlated degrees of freedom (DOFs) in the control policy that requires high dimensional function approximators in the learning agent. For example, to achieve certain contact force in the target direction, the robot can either create a reference position toward the object or adjust its stiffness alone, or regulate both the reference positon and stiffness, which raises a control problem of an over-actuated system. These function approximators with higher dimension increase the sample complexity of the learning algorithm and limit its applicability to physical systems [14].

We propose to address this challenge by developing a method that can utilize the task dynamics to accelerate learning. Model-based RL methods can significantly reduce the required sample time by acquiring the system dynamics and further discovering an effective policy [15]. Though the task dynamics is not fully accessible, it can be approximated with some learned model. The iterative Linear Quadratic Gaussian (iLQG), a Model-Predictive Control (MPC) method, is proposed to optimize trajectories by iteratively constructing locally optimal linear feedback controllers under a local linearization of the dynamic model and a quadratic expansion of the rewards [16]. This method is verified on a stochastic model of the human arm, with 10 state dimensions and 6 muscle actuators [17, 18]. However, with a partial knowledge of the task model and a linear feedback control law, its ability to find optimal strategies is very problematic. To further improve the performance of the learned policy, the learned models and iLQG are combined for accelerating model-free reinforcement learning with a large scale of nonlinear approximators by imagination rollouts [19]. Imagination rollouts (ImR) is a way to accelerate experience collection by generating synthetic on-policy trajectories under a learned model. Adding these synthetic samples (imagination rollouts) to the replay buffer is shown to effectively augment the amount of experience available for the algorithm [19]. With this scheme, however, a model representing a good approximation to the dynamics as well as the rewards must be provided to the algorithms to produce accurate dynamics predictions. Otherwise, the samples of imagination rollouts will diverge from that of actual experience and aggravate the learning process.

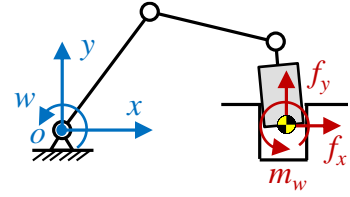

Fig. 1. Considered robotic assembly system. The peg-in-hole problem is implemented in the assembly plane $oxy$. $x$, $y$, $w$ present the positive directions of the robot motion (rotation), while $f_x$, $f_y$, $m_w$ present the positive total external force (moment).

In our previous work, the variable stiffness skill of robot is obtained by reinforcement learning [20]. The specific method we used is DDPG, an actor-critic, model-free algorithm that can operate in continuous action spaces [21], which is especially appealing for learning the interaction strategy between the robot and environments. In this work, the learned model is incorporated into a RL algorithm to generate good exploratory behaviors through trajectory optimization. The iLQG algorithm is utilized to generate good trajectories which are then mixed together with on-policy experience in the replay buffer. The task-relevant samples collected in the model-guided exploration (MGE) provide an initialization to the RL agent of DDPG for learning acceleration. In order to obtain good model-based exploration and improve the efficiency of deep reinforcement learning, we need to find an effective model-learning algorithm to estimate model dynamics. Existing studies have proposed a variety of modeling method, including locally-weighted regression [22], Gaussian processes [23], and neural networks [24]. Instead of learning a global model for all observations and actions, we focus only on a good local model around the latest set of samples and adapt the idea of iteratively refitted time-varying linear models which is proposed in [25] for its simplicity and applicability to nonlinear and discontinuous models. During exploration, the linear dynamics is iteratively fitted by combining the information of a dynamics prior and the robot's recent experience. Generally, there are several possible choices for the dynamics prior including Gaussian priors [19], neural network priors [15], and human demonstration [26]. The Gaussian prior is limited to representing globally linear dynamics and has limited representational capacity in complex state spaces, while the neural network prior must be trained on previous interaction data in order to provide a helpful prior model of the system dynamics. Learning a model prior from human demonstrations is an intuitive scheme and proved efficient in simple manipulation tasks. However, in the problem variable compliance control, the stiffness policy cannot be gathered directly from a human mentor so that it is difficult for implementation. In this work, we use a coarse analytical model with respect to the insertion task as a prior and iteratively update model dynamics with online adaptation through Kalman filter [27]. Compared with the data fusion method based on normal-inverse-Wishart prior in previous studies [15, 25], the Kalman filter with elegant recursive properties is more suitable for incorporating a analytical prior in the estimator [28]. We demonstrate this approach by evaluating its accuracy of model estimation in peg-in-hole experiments.

Another challenge for the skill learning in physical robot is the threat from the large contact force during operation that can cause serious damage to the robot and workpieces. Moreover, as the robot overload may happen several times in a training episode, it tends to cost a lot of time and slow down the learning process if the reaction strategy is not carefully formulated. To address this problem, we propose a safety controller called largest possible potential energy restriction (LPPR) that works parallel to the online task planner. When overload appears, it will compulsively reduce the largest possible potential energy in the corresponding DOF and pull system back to safe region in one step. In addition to restricting contact force, the safety controller can bring the robot a more diverse initial state distribution and therefore promote generalization of the learned policy [29].

The main contribution of this work comes from two complementary techniques for improving the efficiency of compliance skill learning with respect to robotic assembly: (a) a method for combining the deep RL algorithm with analytical priors so as to accelerate learning while preserving the benefits of model-free RL, and (b) a safety reaction strategy for restricting the contact force during exploration. The rest of this paper is structured as follows. Section II firstly formulates the skill learning problem for peg-in-hole task and accordingly proposes the force restriction strategy. Section III briefly reviews the Kalman filter and further explains how it is applied to the learning process of DDPG. Section IV presents the experimental evaluation of our methods on a 7-DOF robot manipulator. Finally, Section V concludes this research and addresses future works.

## II. COMPLIANT CONTROL FOR ROBOTIC ASSEMBLY

### A. Peg-in-hole task

The peg-in-hole problem is generally divided into two phases: a) the search phase and b) insertion phase [30]. In the first phase, the task is to locate the hole by visual or tactile feedback [31]. While this study focused on the more challenging insertion phase where force control is explicitly required. The robot has to align the axis of the peg with respect to that of the hole and pushes the peg to the desired depth. In high precision assembly, even a slight misalignment will lead to high friction and resultant jamming. The clearance between the peg and hole is generally smaller than the uncertainty of the position sensing and control so that neither traditional kinematic planning nor position control can undertake this task.

In most studies, the assembly process is analyzed in decoupled assembly planes which are defined as the two orthogonal planes parallel to the hole axis [2, 3, 32-36]. Without loss of generality, we consider a 2D peg-in-hole problem wherein a square peg and a hole are implemented in the assembly plane $oxy$. A 3-DOF robot manipulator grasps the peg and interacts with the hole. Based on the feedback of contact force (torque) at the peg tip, the robot controller guides the insertion process in Cartesian space (Fig. 1).

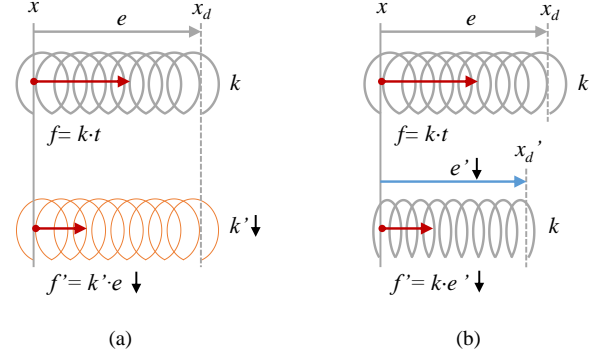

Fig. 2. Two strategies for contact force restriction. The compliant robot is represented by a 1-DOF extension spring with an original length of zero. (a) Decreasing the stiffness from $k$ to $k'$. (b) Decreasing the positional deviation from $t$ to $t'$.

### B. Variable Compliant Control

In order to ensure positioning accuracy and stiffness, traditional industrial robot usually utilizes negative feedback controller with high gains. Unfortunately, high-gain control is not favorable for many manipulations where sophisticated robot-environment interaction is included, e.g. the insertion task. In contrast, impedance control [37] is one of the most popular approaches to Cartesian compliance control [38] and it seeks to maintain a mass-damper-spring relationship between the external force $f$ and position displacement $\Delta x = x_d - x$ in Cartesian space:

$$f = M\ddot{e} + D\dot{e} + Kt \tag{1}$$

where $e = x_d - x$ represents the difference between the actual position $x$ and the moving reference point $x_d$ which is called the *Positional Deviation*. $M, D, K$ are the positive-definite virtual inertia, damping and stiffness matrices. For most insertion tasks, however, the robot is supposed to operate slowly and smoothly during contact so that the operation can be regarded as a quasi-static process. The terms related to acceleration and velocity in (1) are relatively insignificant. Consequently, a stiffness controller instead of the classical impedance controller is compatible for robotic assembly. In a Cartesian plane as shown, the robot is supposed to move to the reference trajectory $x_d = [x_d, y_d, w_d]^T$ with a compliance defined by stiffness $K = \text{diag}(k_x, k_y, k_w)$ and meanwhile constrain the contact force $f = [f_x, f_y, m_w]^T$. In practice, the reference trajectory is usually given by incremental form $\Delta x_d = [\Delta x_d, \Delta y_d, \Delta w_d]^T$ for continues movement while the stiffness can vary in a limited range. The input of the variable compliance robot is a combination of elementary movements and designated stiffness:

$$a = [\Delta x_d, k]^T \tag{2}$$

wherein $k = [k_x, k_y, k_w]^T$ represents the visual Cartesian stiffness of the robot for active compliance. For safety reasons, $a$ is limited to a specific interval during training.

$$\begin{cases} a^{upperLmt} = [[\Delta x_d^{Lmt}, \Delta y_d^{Lmt}, \Delta w_d^{Lmt}]^T, [k_x^{Lmt}, k_y^{Lmt}, k_w^{Lmt}]^T]^T \\ a^{lowerLmt} = [[-\Delta x_d^{Lmt}, -\Delta y_d^{Lmt}, -\Delta w_d^{Lmt}]^T, [0, 0, 0]^T]^T \end{cases} \tag{3}$$

Stiffness selection for insertion are widely discussed in previous researches and one of the desired strategy was proposed in [39] based on model analysis. Although suitable stiffness characteristics seems partly derivable, it is not easy to apply the compliant control to more complex situations such as when significant orientation misalignment or jamming arises.

### C. Strategies for Contact Force Restriction

In high precision assembly, the contact force must be carefully constrained to avoid robot overload and workpiece damage. With respect to the 2-D insertion task, the force limitation is given by

$$\begin{cases} f^{upperLmt} = [f_x^{Lmt}, f_y^{Lmt}, m_w^{Lmt}]^T \\ f^{lowerLmt} = -f^{upperLmt} \end{cases} \tag{4}$$

For industrial robots with position controller, the only possible reaction is to move itself in the opposite direction to the excessive external force for alleviation. However, in lack of the contact stiffness and other knowledge of the environment, it is difficult to determine the magnitude of the displacement, so that any active movement is at the risk of leading to a more severe situation. On the other hand, the problem becomes simple when the robot is built with a variable compliance controller due to the extra control variables from the robot stiffness. As shown in the 1-DOF system in Fig. 2 (a), an intuitive strategy is to decrease the stiffness in the threatened direction from $k$ to $k'$ and make the robot more compliant to the external force. Though this passive compliance is able to ease the immediate tension effectively, it cannot take the robot away from the threat that fundamentally comes from the large positional deviation. In the following exploration, the robot will cause overload once again when the stiffness rises. In order to take the robot back to the safe region in favor of further learning, we choose to reduce the positional deviation by moving the reference position towards the actual position (Fig. 2 (b)).

$$x_d' = \varepsilon x_d + (1-\varepsilon)x \tag{5}$$

Where $\varepsilon \in [0,1]$ is defined as the *Reduction Factor*. Actually, this process has an explicit physical explanation. As the magnitude of the variable compliance robot input is limited, the largest possible potential energy of the spring-like robot system can be scaled by the square of the reduction factor:

$$\max(p') = \varepsilon^2 \max(p), \quad \max(p) = \frac{1}{2}k^{Lmt}e^2. \tag{6}$$

The principle of determination of $\varepsilon$ is to make it a negative feedback to the contact force. Here we choose a simple control law for safety reaction as:

$$\varepsilon = \min\left(1, \left(\frac{f^{Lmt}}{f}\right)^2\right) \tag{7}$$

When the contact force overloads the robot, the LPPR controller will be activated to reduce the positional deviation and keep the system away from danger.

### III. SKILL LEARNING WITH MODEL-BASED ACCELERATION

#### A. A RL-based controller for insertion task

We utilize a learning-based controller (LBC) working in the level higher than the variable compliance robot to find optimal policies through trial and error for the insertion task. It is supposed to map states $s \in \mathbb{R}^{D_s}$ observed from environments to a specific robot action $a \in \mathbb{R}^{D_a}$ with deterministic reference trajectory and stiffness. Firstly, we need to design observation states according to the assembly process as with feedback signals for a controller system. In time step $t$, the controller observes the current state $s_t$ of the system and generate action $a_t$ to react to this observation. The definition of $s_t$ is partially inspired by the Proportional-differential controller:

$$s_t = [e_t, f_t, \Delta f_t, \Delta x_t, \xi_t]^T \tag{8}$$

where $\Delta x_t = x_t - x_{t-1}$ and $\Delta f_t = f_t - f_{t-1}$ are the incremental displacement and incremental force respectively. While the differential signals in PD controller are included to ensure the system stability, $\Delta x_t$ and $\Delta f_t$ are used to infer if the peg is jammed in a certain direction. $e_t = [e_{x,t}, e_{y,t}, e_{w,t}]^T$ represents the positional deviation of all the three DOFs. Instead of explicit position information, we include the percentage of insertion $\xi_t = d_t/d$ in the state vector, where $d_t$ and $d$ are the current insertion depth and the total depth of the hole respectively. The action $a_t$ generated by the LBC is defined as (2). In order to explore new control policies, we add exploration noise $\delta \in \mathbb{R}^{D_a}$ to the action

$$a_t = a_t + \delta \tag{9}$$

After the execution of $a_t$, a reward $r_t \in \mathbb{R}$ from the environment is provided to the controller for parameter updating and the system moves to a next state $s_{t+1}$. In the implementation, both $s$ and $a$ are normalized in the preprocess stage to give a better learning performance. According to the expected behavior of the robot in the insertion task, we define the reward function as follows:

$$r_t(a_t, s_{t+1}) = -w_{dis}\frac{\Delta y_{t+1}}{\Delta y_d^{Lmt}} - w_f\left(\left\|\frac{f_{t+1}}{f^{upperLmt}}\right\|_\infty\right)^2$$
$$- w_{stf}\left(\left\|\frac{k_t}{k^{upperLmt}}\right\|_\infty\right)^2 + r_{end} \tag{10}$$

The first three items represent the immediate rewards in each step that are respectively used to punish the displacement away from the hole, large contact force and high stiffness. We can regulate the relative weights of these terms by adjusting

$w_{dis}, w_f, w_{stf}$. The starting point of the reward design is to encourage downward movement of the peg and meanwhile discourage actions that will cause large assembly forces or require high stiffness motion. If the trial goes to the termination state before reaching the maximum number of step, $r_{end}$ will be added to the reward in the terminal state of current episode.

$$\begin{cases} r_{end} = 1, \ \xi > 0.98 \\ r_{end} = -1, \ f \notin [f^{lowerLmt}, f^{upperLmt}] \ . \\ r_{end} = 0, \ \text{else}. \end{cases} \quad (11)$$

The termination conditions help with reward shaping by penalizing sequences of action that lead the agent to undesirable parts of the state space. It is equal to a pseudo absorbing state when the system moves outside the allowable state space [29]. The total cost to be minimized is the sum of $r_t$ in the finite horizon $r_{total} = \sum_{t=1}^{T} r_t$.

In order to training the proposed controller online in real robot system, we need a stable and efficient algorithm. Here the deep RL algorithm of DDPG is utilized for its sampling efficiency. DDPG integrates a replay buffer to store transitions $(s_i, a_i, r_i, s_{i+1})$ sampled from the training process and update the actor and critic at each timestep by sampling a minibatch uniformly from the buffer. A neural network termed as *actor* specifying the current variable compliance policy implements the LBC that is represented as $a_i = \pi^{LBC}(s_i)$. For updating the controller, a *critic* network is used to evaluate and update the actor.

### B. Model-Guided Exploration

One natural approach to incorporating model information into an RL algorithm as DDPG is to employ a learned model to generate good exploratory behaviors through planning or trajectory optimization. In this work, we use the iLQG to optimize the actions with respect to the cost given in (10) and (11), under an estimated dynamics model called *LocalModel*. This is realized by re-optimizing the state sequence and associated action sequence at each time step of the control loop, which is starting at the current state. The procedure is repeated so that the state sequence being optimized extends to some pre-defined horizon. For exploration, we choose a short horizon to reduce the amount of computation at the cost of some myopic behaviors. Details about the iLQG algorithm can be found in [16]. In our experiment, a one-step iLQG optimizer is used to implement the exploration policy $a_i = \pi^{MPC}(s_i)$.

Before the LBC starting to work, the robot uses MPC to generate actions under the model, and then collect experiences along these trajectories by appending these to the replay buffer. These experiences are further used to give primary training of the learning agent to achieve a warm start of the DDPG controller. The estimated dynamics model is linear but refitted at each time step based on the recently observed states and actions, as well as the dynamics prior.

TABLE I
PREDICTION PERFORMANCE WITH DIFFERENT METHOD

| method | winning rate | mean error | error std |
|---|---|---|---|
| Prior | 0.008 | 0.1882 | 0.1051 |
| Linear | 0.015 | 0.2537 | 0.2399 |
| Average | 0.131 | 0.1772 | 0.1357 |
| Kalman filter | 0.846 | 0.1513 | 0.1014 |

### C. Fitting Dynamics with Priors through Kalman filter

This section describes how the linear dynamics can be fitted under a dynamics prior, as well as the scheme for updating the dynamics online based on the robot's recent experience. In order to fit a linear model to a set of $n$ samples $\{y_1, y_2\}$, where $y_1 = [s_i, a_i]^T$, $y_2 = s_{i+1}$, and $i = t-n, ..., t-1$, we can simply use a multi-variable Gaussian model to obtain $p(s_{t+1} | s_t, a_t)$. Let $z_{t+1}$ and $R_{t+1}$ be the empirical mean and the empirical covariance of the estimated future state, respectively.

$$\begin{aligned} z_{t+1}(s_t, a_t) &= \mu_2 + \Sigma_{21} \Sigma_{11}^{-1} ([s_t, a_t]^T - \mu_1) \\ R_{t+1} &= \Sigma_{22} - \Sigma_{21} \Sigma_{11}^{-1} \Sigma_{12} \end{aligned} \quad (12)$$

Wherein $\mu$ is the average vector of the dataset $\{y_1, y_2\}$, and $\Sigma$ is the covariance matrix with respect to $y_1$ and $y_2$. As discussed in the previous section, the online estimate of the dynamics linearization makes use of a dynamics prior. In the control problem of variable compliance manipulation, though it is not well defined, we have some prior knowledge of the model dynamics in analytical form. In general, the state is estimated based on the assumption that the end effector of the robot is clamped by the environment, which is often the case during insertion, and makes little movement. Thus, the future state $\hat{s}_{t+1}$ is estimated according to the decoupled spring model as shown in Fig. 2.

$$\begin{cases} \hat{e}_{t+1} = \text{diag}(\eta) \ (e_t + \Delta x_{d,t}) \\ \hat{f}_{t+1} = \text{diag}(k_t) \ \hat{e}_{t+1} \\ \Delta \hat{f}_{t+1} = \hat{f}_{t+1} - f_t \\ \Delta \hat{x}_{t+1} = (I - \text{diag}(\eta)) \ (e_t + \Delta x_{d,t}) \\ \hat{\xi}_{t+1} = \xi_t - \Delta \hat{y}_{t+1} / d \end{cases} \quad (13)$$

Wherein $\eta = [\eta_x, \eta_y, \eta_w]^T$ represents the probability expectation for complete clamping in *x*, *y*, *w* directions respectively. For example, $\eta_y = 1$ means that the robot is supposed to be completely constraint by the environment at all the time and will not make any movement in *y*-direction. While $\eta_y = 0$ implies that the robot is moving in free space and perfectly follows its reference trajectory. It is obvious that both the above situations cannot characterize the actual peg-in-hole process. In fact, $\eta$ is a state variable of the dynamics model that varies from 0 to 1. In this research, we are not going to find a precise mathematical model, and $\eta$ is considered as a constant

**Algorithm 1** Deep RL with MGE and LPPR

---

Initialize DDPG networks with the random parameters.
Initialize replay buffer $R_{DDPG}$ (size $b$, FIFO).
Initialize the linear fitting buffer $R_{fitting}$ (size $n$, FIFO)
**for** *episode* = 1 **to** $M$ **do**
  Reset the environment and receive initial observation $s_1$.
  **for** $t$ = 1 **to** $T$ **do**
    **if** *episode* <= $I$ **then**
      $a_t \leftarrow \pi^{MPC}(s_t)$.
    **else**
      $a_t \leftarrow \pi^{LBC}(s_t)$
    **end if**
    Add random noise to $a_t$.
    $\varepsilon_t \leftarrow lppr(s_t)$
    $u_t \leftarrow (a_t, \varepsilon_t)$
    Execute the $u_t$ and receive reward $r_t$; get next state $s_{t+1}$.
    **if** *lppr* is inactivated **then**
      Store transition $(s_t, a_t, r_t, s_{t+1})$ in $R_{DDPG}$
      Store transition $(s_t, a_t, s_{t+1})$ in $R_{fitting}$.
    **end if**
    Update DDPG networks and $\pi^{LBC}$.
    **if** $R_{fitting}$ is full **then**
      *LocalModel* ← FitDynamics($R_{fitting}$) via (12)(14)(17)
      $\pi^{MPC}$ ← iLQG_OneStep(*LocalModel*)
    **end if**
  **end for**
  $R_{fitting} \leftarrow \varnothing$
**end for**

---

parameter. Instead, we use (13) as the initial knowledge of the system in a Kalman filter.

The Kalman filter is one of the most important and common data fusion algorithms in use today. In light of its idea, the best estimate we can make of the state of the system is provided by combining the knowledge from the prior model and the online measurement. This fusion algorithm involves two stages: model-based prediction and online update. Assuming that there does not exist noise in the control inputs of MPC, the Kalman filter equations for the prediction stage are

$$\hat{s}_{t+1|t} = F_t s_t + B_t a_t$$
$$P_{t+1|t} = F_t P_{t|t} F_t^T \tag{14}$$

where

$$F_t = \begin{bmatrix} \mathrm{diag}(\eta) & 0 & 0 & 0 & 0 \\ \mathrm{diag}(\eta \circ k_t) & 0 & 0 & 0 & 0 \\ \mathrm{diag}(\eta \circ k_t) & -I & 0 & 0 & 0 \\ I - \mathrm{diag}(\eta) & 0 & 0 & 0 & 0 \\ [0,1,0](\mathrm{diag}(\eta)-I) & 0 & 0 & 0 & 1 \end{bmatrix} \tag{15}$$

$$B_t = \begin{bmatrix} \mathrm{diag}(\eta) & 0 \\ \mathrm{diag}(\eta \circ k_t) & 0 \\ \mathrm{diag}(\eta \circ k_t) & 0 \\ I - \mathrm{diag}(\eta) & 0 \\ [0,1,0](\mathrm{diag}(\eta)-I) & 0 \end{bmatrix}^T \tag{16}$$

are respectively the state transition matrix and control input matrix corresponding to (13). $P_{t|t} = \mathrm{E}[(s_t - \hat{s}_{t|t-1})(s_t - \hat{s}_{t|t-1})^T]$ is the covariance matrix of the estimation error in time step $t$ that can be approximated by calculating the average error covariance from a number of previous steps $(s_{i+1}, \hat{s}_{i+1|i})|_{i=t-n,\ldots,t-1}$. Then the prediction is fused to the empirical information from the local dynamics fitting. The posteriori estimates are given by

$$\hat{s}_{t+1} = \hat{s}_{t+1|t} + K_t(z_{t+1} - \hat{s}_{t+1|t})$$
$$P_{t+1} = P_{t+1|t} - K_t P_{t+1|t} \tag{17}$$

where $K_t$ is the Kalman gain matrix, and it is calculated as

$$K_t = P_{t+1|t}(P_{t+1|t} + R_{t+1})^{-1} \tag{18}$$

The Kalman gain is adaptively adjusted based on the relative accuracy of the empirical and prior dynamics estimates. The intuition behind is that, when the prior is less effective, we should weight the empirical estimate more highly. The equations of (12)(14)(17) are combined to build a local model of the system dynamics $\hat{s}_{i+1} = LocalModel(s_i, a_i)$. Note that in the proposed modeling method, which will be called *Kalman filter-based Local Model Fitting* (KLMF) in the following sections, the only hyper parameter to be determined is the clamping probability $\eta$, which we set as $\eta = [0.99, 0.99, 0.99]^T$ in the following experiments.

The fusion method under the normal-inverse-Wishart prior that utilizes a constant weighing parameter for balancing the information from online fitting and priors shows little adaptability. In most cases, it is equivalent to a weighted average synthesizer in practice [15, 19, 25]. To illustrate the advantages of Kalman filter in state prediction for MPC, we compare its estimation accuracy to that of prior information, local linear fitting, and weighted average synthesis (Table. 1). The database used for evaluation is collected from the actual system in previous experiments and compose of 1000 $(s_i, a_i, s_{i+1})$ tuples. The prediction error is defined as $\|\hat{s}_{i+1} - s_{i+1}\|_2$. The winning rate of each method shows its frequency of giving the best estimation over others in the dataset. From the result, it is obvious that the Karman filter is able to exploit the data from the analytical prior and local linear fitting effectively to generate a more precise prediction of the future state, which is outperform the existing average synthesis method.

*D. Algorithm Summary*

The deep RL algorithm with the proposed model-guided exploration and contact force restriction strategy summarized in Algorithm 1. The main difference of our algorithm from the original DDPG is that the state-action controller is assumed by the MPC instead of LBC in the early stage to achieve better exploratory behaviors. At each time step, the algorithm observes the current state $s_{t+1}$ and adds $(s_t, a_t, s_{t+1})$ to the finite-length buffer $R_{fitting}$ to update the empirical mean $z_t$ and covariance $R_t$ of the linear model. The empirical (12) and prior estimates (14) are then combined in a Kalman filter (17) to construct the posterior mean and covariance, from which we can obtain a *LocalModel*. This local model is used, together with the cost function (10), to optimize a new MPC policy $\pi^{MPC}$ using iLQG. Then we use this MPC policy to choose the action for exploration. In addition to the optimal trajectories, it is desirable to collect diverse experience that contains both good and bad actions for a good learning process [19]. To that end, we add a small amount of Gaussian noise to the action.

TABLE I
SPECIFIC PARAMETERS OF LBC

| Parameter | Value |
|---|---|
| $(f_x^{Lmt}, f_y^{Lmt}, m_w^{Lmt})$ | (40 N, 40N, 5 Nm) |
| $(\Delta x_d^{Lmt}, \Delta y_d^{Lmt}, \Delta w_d^{Lmt})$ | (1 mm, 1 mm, 2°) |
| $(k_x^{Lmt}, k_y^{Lmt}, k_w^{Lmt})$ | (4000 N/m, 4000 N/m, 200 Nm/rad) |
| $(w_{dis}, w_f, w_{stf})$ | (1, 1, 0.1) |

During learning trails, the LPPR controller works side by side with the state-action controller, and generates reduction factor in directions of $x$, $y$, and $w$

$$\boldsymbol{\varepsilon} = [\varepsilon_x, \varepsilon_y, \varepsilon_w]^T \leftarrow lppr(\boldsymbol{s}) \quad (19)$$

that is incorporated into the extended action defined as

$$\boldsymbol{u} = [\boldsymbol{a}, \boldsymbol{\varepsilon}]^T \quad (20)$$

It is the extended action $\boldsymbol{u}$ that effects the next state of the robot-environment system. For MPC or LBC, $\boldsymbol{u}$ is not a controllable variable, so that the step with safety reaction is beyond their regulation and should not be used for updating. So that the experience transitions are added to the training buffers only when the LPPR controller is inactivated.

## IV. EXPERIMENTAL EVALUATION

### A. Experimental Setup

We evaluate our method on a peg-in-hole task using the a 7-DOF torque-controlled robot manipulator. The robot is shown in Fig. 4 together with the peg-in-hole environment. Each joint of the robot is integrated with a torque sensor so that the controller is able to estimate Cartesian contact force $(f_x, f_y, m_w)$ with force Jacobian [40]. Similarly, the position and orientation of the peg $(x, y, w)$ is calculated though forward kinematics. To realize variable compliance control, we utilize a Cartesian impedance controller without inertia shaping [37]. The position error of the robot is about ±0.5 mm in translation and ±0.5° in rotation. The 2D hole is implemented by an adjustable slot with two parallel walls perpendicular to the assembly plane *oxy* (Fig. 4). Our hole and peg are both made of steel with different sizes. Their diameters (or width) are D=23.04 mm and d=23 mm respectively. The depth of the hole is H=36 mm. Note that the peg is machined with a fillet of 0.5 mm to facilitate the initial location of the peg and meanwhile avoid serious scratch to the hole. The proposed algorithm is implemented on a PC (CPU 3.5 GHz, RAM 32GB) communicating with the robot controller by Socket. The PC receives observation from the robot, trains the networks, and sends action command to the robot controller in every 500 ms. This cycle time is selected though trails in experiments to ensure a quasi-static process.

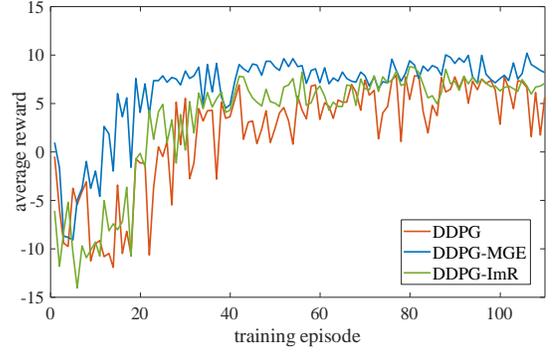

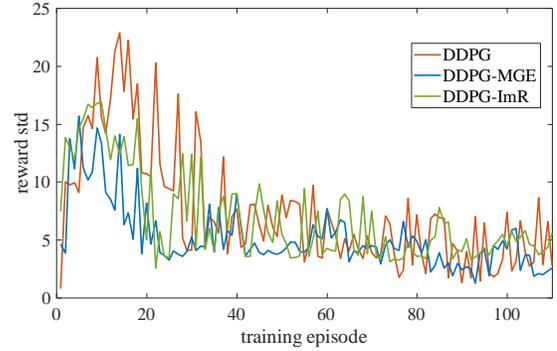

Fig. 4. Learning curves of three strategies. (a) Average reward of 10 trials. (b) Standard deviation of reward.

For the RL algorithm, the neural networks of the actor and critic of DDPG have same structures of two fully connected hidden layers with 64 units. Adam algorithm [41] is used to update the networks with learning rates of $3\times10^{-4}$ and $1\times10^{-4}$ respectively for the actor and critic. The memory capacity of the replay buffer is $b = 9000$ and the size of the minibatch is 100. For MPC, the buffer used to fit linear model has a capacity of $n = 5$. A list of specific parameters of the learning-based controller is shown in Table. 1.

### B. Comparisons in learning speed

We evaluated the model-guided exploration method in comparison to another model-based acceleration method with imagination rollouts as well as the original DDPG. Both the model-based methods utilize the local dynamics estimated by KLMF. In the training phase, the robot starts movement from a position with an initial angle error $e_w \sim U(-3°, 3°)$ as shown is Fig. 3. The derivation in starting orientation gives robot the chance to learn the strategy in different situations and further acquire a general skill. For action exploration in each step, we use noise sampled from a normal distribution with standard deviation of 10% of the action range. In order to eliminate random disturbance with respect to the peg-in-hole task, each method is tested in skill learning runs for 10 times. Note that each of the runs is performed separately with no information retained between runs. The evaluation is given in average reward and standard deviation of the reward as shown in Fig. 4.

TABLE II
SUCCESS RATE WITH DIFFERENT INITIAL ORIENTATION

| $e_w$ | DDPG-MGE | DDPG-ImR | DDPG |
|---|---|---|---|
| 0° | 1.0 | 1.0 | 1.0 |
| 1° | 1.0 | 1.0 | 1.0 |
| 2° | 1.0 | 0.9 | 1.0 |
| 3° | 1.0 | 0.9 | 0.9 |
| 4° | 0.9 | 0.8 | 0.8 |
| 5° | 0.8 | 0.5 | 0.6 |
| 6° | 0.5 | 0.2 | 0.2 |
| 7° | 0.5 | 0.1 | 0.0 |
| 8° | 0.2 | 0.0 | - |
| 9° | 0.0 | - | - |

After 100 training episodes (about $1\times10^4$ steps), all the learning strategies shows a convergence in task reward. The learning curves illustrate the effect of applying MGE and ImR to the peg-in-hole task. It is noticeable that mixing the good estimation of the task model can generally improve the learning speed (Fig. 4 (a)) and results in a more consistent convergence process (Fig. 4 (b)). On the other hand, it shows that mixing imagination rollouts to the replay buffer does not significantly improve data-efficiency, while model-guided exploration is able to speed up the convergence experience. This observation seems inconsistent with the results of [19] concluding that ImR is more efficient than MGE. However, it reveals the essential difference between our work and previous studies in RL theories.

1) Model uncertainty. In this work, the problem to be solved is a highly uncertain physical system instead of a simulator, so that neither the ground truth dynamics nor the cost function with an idea form is available for the model-based acceleration method. In the scheme of imagination rollouts, the nonideal model and reward is prone to produce synthetic experience with low quality that can potentially mislead the learning direction. By contrast, the model-guided exploration is only used to generate more efficient exploration trajectories than that generated by LBC with newly initialized parameters, during which the collected transitions are guaranteed to reflect the true environment. In other words, the MGE method is able to accelerate the learning process while reserve the model independence property of the RL algorithms.

2) High-dimensional task space. As shown in the introduction, the skill-learning problem with variable compliance control features a large number of correlated DOFs in action space. Both of the reference positon and the stiffness in each DOF of the robot are supposed to be optimized for the task. With a model-free RL algorithm, it can take a long time to collect enough data and then figure out the relationship between inputs and outputs of the over-actuated system. However, MPC is able to use the covariance of state-action pairs to guide the exploration. This covariance information comes from the local model given by KLMF, which combines the description from

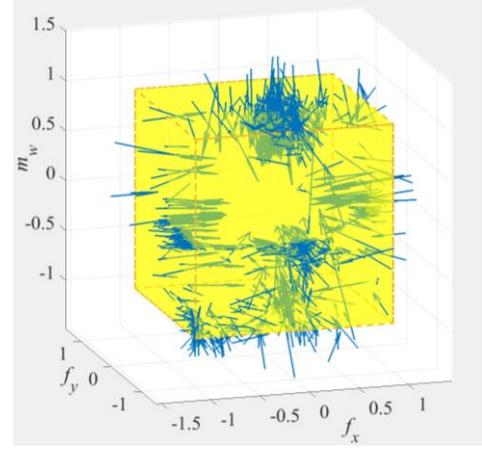

(a)

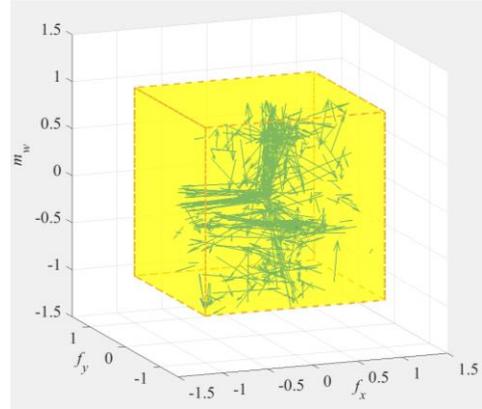

(b)

Fig. 5. Changes in contact forces in training steps. (a) The contact forces decrease to legitimate range under LPPR controller. (b) In the next step, the contact forces change within the legitimate area.

(12) and (13) in an optimal way. Consequently, the transitions collected under MPC are more relevant to the task and can be used to train the learning agent more efficiently than the original RL setup.

C. *Comparisons in policy robustness*

For comparisons in learning speed, each learning strategy is used to give 10 training runs and finally we have 10 instances of the trained RL controller for each strategy. In this section, the robustness of these controllers to initial orientation error in the insertion task are tested. The orientation error $e_w$ varies from 0° to 9° with an increment of 1° in the experiment. For each $e_w$, we evaluate the performance of the learning strategy in 10 trials with different controller instance and record its success rate. The results are given in Table II. It is clear that the variable compliance policy learned by DDPG strengthened by MGE has strong generalization ability for situations beyond the training set. It significantly outperforms the ImR method and the original DDPG in terms of final performance.

D. *Contact force restriction*

Proper contact force restriction is compulsive for a safe and coherent learning process in physical robots. In our experiments, the proposed LPPR method is used to deal with robot overload. During the training of DDPG with MGE (10 trials, or 1100 episodes), the robot overrun the force limitation for 816 times in total. In Fig. 5, the legitimate area for contact force is given as a yellow cubic, and the force change between two consecutive observations is represented as blue vectors pointing from the starting state to the terminal state. Except for the opening direction of the hole (positive *y*-direction), the overload threat comes from all the rest directions. Each time the overload appears, the LPPR controller is activated and decrease the contact force to the legitimate range (Fig. 5 (a)). Fig. 5 (b) shows that the attenuation of the largest possible potential energy of the system is enough to ensure that the robot keeps away from overload and will stay within the safe area in the next step of operation.

## V. CONCLUSION

In this work, we presented a model-guided exploration (MGE) strategy to accelerate the skill learning for variable compliance robot. It characterizes a Kalman filter to fuse the data from the adaptive linear dynamics and a coarse prior model from analytical description, which we call Kalman filter-based local model fitting (KLMF). It is the dynamics model estimated by KLMF that is incorporated into MPC to choose the actions for efficient exploration. This allows us to exploit available model knowledge for decreasing sample complexity, while preserving the benefits of model-free reinforcement learning. In the experiments with a 7-DOF robot manipulator, we show that, in comparison to recently proposed method based on synthetic on-policy rollouts and the original RL methods, our scheme tends to learn faster and acquires more robust policies. We further explore how the contact force during robot manipulation can be constrained in the framework of variable compliance control, without interfering the learning process. We show that the proposed scheme based on largest possible potential energy restriction (LPPR) is able to effectively eliminate the overload and create a safe initial condition for the next operation.

The two innovations of the proposed strategy, local model fitting based on Kalman filter for model-guided exploration and LPPR for safety reaction, significantly improves the learning efficiency of deep RL algorithm. While the application of the proposed strategy is confined to robotic assembly, it can be easily generalized to a variety of variable compliance manipulations with coarse analytical priors, which is exactly our future work.

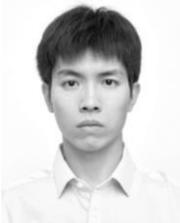

**Tianyu Ren** received the B.S. degree in mechanical engineering from Beijing Institute of Technology, Beijing, China, in 2014. He is currently pursuing the Ph.D. degree at the department of mechanical engineering, Tsinghua University, Beijing, China. His research focuses mainly on robot force control and manipulation.

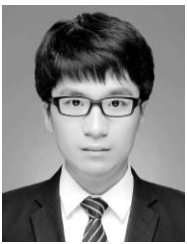

**Yunfei Dong** received the B.S. degree in School of Aerospace, Tsinghua University, Beijing, China. He is currently working towards the Ph.D. degree at Department of Mechanical Engineering, Tsinghua University, Beijing, China. His research interests focus on robot dynamics and control.

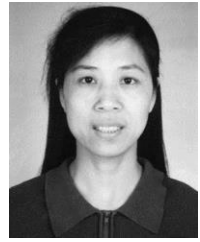

**an Wu** received the B.S., M.S., and Ph.D. degrees in mechanical engineering from Tsinghua University, Beijing, China, in 1988, 1990, and 2008, respectively. She is currently a professor in the Department of Mechanical Engineering, Tsinghua University, Beijing, China and she is a member of American Society for Precision Engineering, and a senior member of the Chinese Mechanical Engineering Society.

Her research interests focus on precision and ultra-precision machining, and micro-manipulator for life science. She has authored/coauthored more than 120 research papers, five book chapters, and over 20 granted patents. She won the Science and Technology Progress Award from Ministry of Education of China and the Award of Beijing Higher Education Achievements.

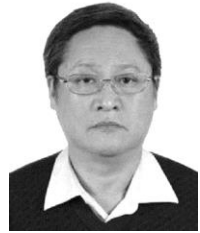

**Ken Chen** received the B.S. degree in mechanical engineering from Sichuan University, Chengdu, China, in 1982, and the M.S. and Ph.D. degrees in mechanical engineering from Zhejiang University, Hangzhou, China, in 1984 and 1987, respectively. He is currently a professor in the Department of Mechanical Engineering, Tsinghua University, Beijing, China. From 1991 to 1992, he was a Visiting Professor at the University of Illinois, Chicago. From 1992 to 1995, he was a Postdoctoral Researcher with Purdue University, Indianapolis. He has published more than 200 papers. His research interests include bionic-robots, special robots and robotic manufacturing. Prof. Chen is a member of the American Society of Mechanical Engineers, and a senior member of the Chinese Mechanical Engineering Society.